\pdfoutput=1

\documentclass[11pt]{article}

\usepackage[]{acl}

\usepackage{times}
\usepackage{latexsym}

\usepackage[T1]{fontenc}

\usepackage[utf8]{inputenc}
\usepackage{subcaption}
\usepackage{graphicx}
\usepackage{balance}

\usepackage{microtype}

\title{Vocabulary Transfer for Biomedical Texts: \\
Add Tokens if You Can Not Add Data}

\author{Priyanka Singh \\
Higher School of Economics\\
St. Petersburg, Russia\\
\And
Vladislav Mosin\\
LEYA lab,\\
Higher School of Economics\\
St. Petersburg, Russia\\
\And
Ivan P. Yamshchikov\\
CAIRO, THWS\\
W\"{u}rzburg, Germany\\
\texttt{ivan.yamshchikov@thws.de}
}

\begin{document}
\maketitle
\begin{abstract}

Working within specific NLP subdomains presents significant challenges, primarily due to a persistent deficit of data. Stringent privacy concerns and limited data accessibility often drive this shortage. Additionally, the medical domain demands high accuracy, where even marginal improvements in model performance can have profound impacts. In this study, we investigate the potential of vocabulary transfer to enhance model performance in biomedical NLP tasks. Specifically, we focus on vocabulary extension, a technique that involves expanding the target vocabulary to incorporate domain-specific biomedical terms. Our findings demonstrate that vocabulary extension, leads to measurable improvements in both downstream model performance and inference time.


\end{abstract}
\section{Introduction}

 The Transformer architecture, introduced by \citet{vaswani2017attention}, has revolutionized natural language processing across various domains. In the biomedical field, several Transformer-based models have been specifically tailored for biomedical corpora, including BioBERT \cite{lee2020biobert}, PubMedBERT \cite{gu2021domain}, ClinicalBERT \cite{Huang2019ClinicalBERT}, and MedGPT \cite{Zuo2019MedGPT}, among others. The complexity of tokenization in biomedical texts arises from multiple factors. Biomedical language often diverges significantly from general English in both syntax and lexicon, frequently incorporating complex compound terms, non-standard abbreviations, and specialized terminologies that reflect the field's dynamic and rapidly evolving nature. Biomedical literature often includes acronyms, abbreviations, digits, internal capitalization, special characters, and structured information like medical codes and timestamps.

This paper explores applicability of vocabulary transfer introduced in \citet{mosin2023fine} to biomedical domain. Traditionally, language models employ the same tokenization method during both initial training and subsequent fine-tuning, typically encompassing thousands of tokens. These tokens can vary from subword units to full words. However, \citet{mosin2023fine} propose that developing a new, task-specific tokenization strategy during the fine-tuning stage may significantly improve model performance.

Vocabulary transfer becomes particularly particularly advantageous when the fine-tuning dataset differs substantially from the one used in initial training. Research in biomedical tokenization has highlighted several problematic cases that exemplify the challenges inherent in this domain, for detailed examples see \citet{Lopez2015}. 



Though in this paper we experiment with biomedical data we believe that similar vocabulary extension approach would be beneficial for other NLP domains where the data is scarce. \cite{gee2022fast} has demonstrated the vocabulary transfer could be beneficial for model compression in business applications. \cite{yamshchikov2022bert} has shown the benefits of vocabulary transfer when using the model trained on the modern Greek texts on the historical texts in ancient Greek, while \cite{remy2024trans,alexandrov2024mitigating} show its benefits when working with low-resource languages. This paper demonstrates that vocabulary transfer is applicable to biomedical texts and can bring certain benefits. We also demonstrate that increasing vocabulary size, i.e. {\em vocabulary extension}, during vocabulary transfer significantly improves downstream performance. We believe this result is not limited to biomedical data but would hold on any other specific domain.
\section{Vocabulary transfer}

In their study, \citet{mosin2023fine} introduce the concept of {\em vocabulary transfer}. Let $V$ denote the original vocabulary obtained during the pretraining phase, comprising $M$ tokens denoted as $\{t_k, v_k\}$, where $t_k$ is a text segment forming a token, and $v_k$ is the corresponding embedding for that token. Then $\widetilde{V}$ represents the new vocabulary utilized during fine-tuning, consisting of $N$ tokens denoted as $\{\widetilde{t}_k, \widetilde{v}_k\}$, where $\widetilde{t}_k$ is a text segment forming a new token, and $\widetilde{v}_k$ is its corresponding embedding. This customized tokenization strategy in the fine-tuning phase facilitates improved model performance on specific tasks or datasets.

To transfer pretrained knowledge from existing tokens to new, corpus-specific tokens, a heuristic token-matching procedure can be employed. In this paper, we evaluate two token initialization heuristics. First, if a token in the new vocabulary directly matches a token in the original vocabulary, its corresponding embedding is assigned to the new token. We refer to this approach as \textit{matched} vocabulary transfer. Additionally, some new tokens may be decomposable into partitions of multiple tokens from the original vocabulary. For each such token in the new vocabulary, we generate all possible partitions comprising tokens from the original vocabulary and select the partition with the minimal number of tokens. If multiple partitions have the same number of tokens, we choose the one containing the longest token. The embedding for the new token is then initialized by averaging the embeddings of the tokens in the selected partition. We refer to this approach as \textit{averaged} transfer. These methods correspond to those described by \citet{mosin2023fine}, where \textit{matched} aligns with the "Match Old Tokens" strategy and \textit{averaged} corresponds to "VIPI".


\section{Data}

We conduct experiments on text classification as a downstream task within the biological domain \cite{mujtaba2019clinical, gao2021limitations, hughes2017medical} using two datasets: OHSUMED \cite{hersh1994ohsumed}, a medical dataset for the classification of cardiovascular diseases, and the Kaggle Medical Texts Dataset\footnote{https://www.kaggle.com/chaitanyakck/medical-text}, which classifies various patient conditions including digestive system diseases, cardiovascular diseases, neoplasms, nervous system diseases, and general pathological conditions. The downstream dataset was split into 80$\%$, 10$\%$ dev, and 10$\%$ test. Our results demonstrate that increasing the number of tokens enhances classifier accuracy when using masked language modeling (MLM) and vocabulary transfer before downstream classification tasks.

Table \ref{tab:data} summarizes the parameters of the datasets that we experiment with.

\begin{table}[h]
\centering
\begin{tabular}{lrr}
 Dataset & Numer of  & Number of  \\
 & Records & Labels\\
\hline
OHSUMED & 13 929 & 23 \\
Kaggle & 28 880 & 5 \\
\hline
\end{tabular}
\caption{Parameters of the datasets used for experiments with the number of records and the number of labels.}
\label{tab:data}
\end{table}

\begin{figure}[h]
		\centering
    	\includegraphics[scale=0.4]{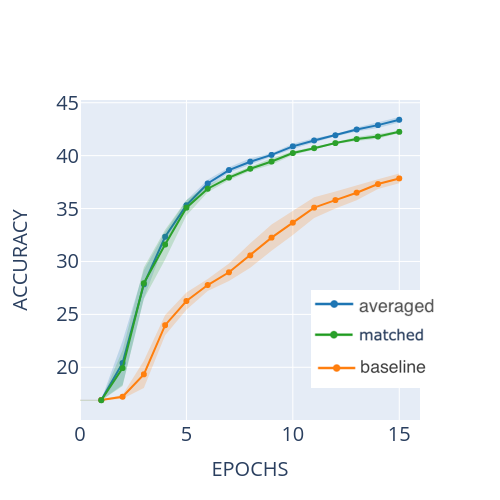}
    	\caption{Performance on Ohsumed data, vocabulary size is 32 000}
    	\label{fig:len_lr}
\end{figure}

\begin{figure}[h]
		\centering
    	\includegraphics[scale=0.4]{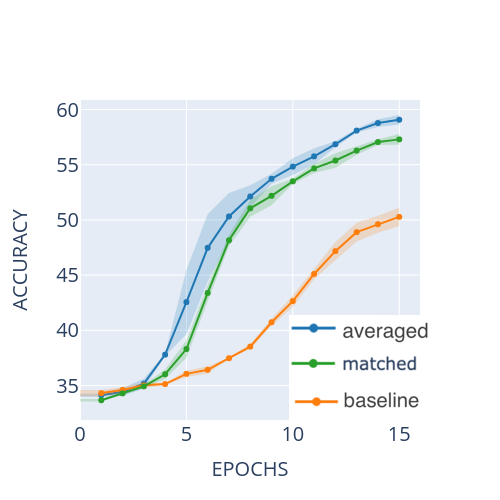}    
    	\caption{Performance on Kaggle Medical dataset, vocabulary size is 32 000}
    	\label{fig:kmd}
\end{figure}

Table \ref{tab:label table} summarizes the total number of frequencies in each label in Kaggle dataset.
\begin{table}[htbp]
\centering 

\begin{tabular}{c| c c c c c}
    \hline
    \textbf{Label} & 1 & 2 & 3 & 4 & 5 \\
    \hline
    \textbf{Frequency} & 3163 & 1494 & 1925 & 3051 & 4805\\
    \hline
\end{tabular}
\caption{Number of Text Data in each Label of Kaggle dataset }
\label{tab:label table}
\end{table}

\section{Experiments}

We conducted a series of experiments using the base version of the BERT model on various medical datasets. Initially, we explored the adoption of a new, dataset-specific vocabulary through an intermediary masked language modeling (MLM) step, involving pretraining on the downstream dataset with updated tokenization. Subsequently, we performed multiple experiments with different parameters and compared the results to a baseline approach, which involved simple fine-tuning on the downstream dataset without applying vocabulary transfer or altering the initial tokenization. This baseline serves as a reference point in our experimental analysis.


\subsection{Fine-tuning Transferred Vocabulary and Change in Classifier Accuracy}

In these experiments, we investigated the limitations of simple token matching in vocabulary transfer and assessed the impact of masked language modeling (MLM) step on the overall performance after vocabulary transfer. We also experiment with the size of the final vocabulary.

First, merely assigning new embeddings to tokens is insufficient for enhancing model performance. Table \ref{tab:classifier_accuracy} presents the relative change in accuracy of downstream classifiers for a medical dataset with five classes without intermediate MLM step and with it. It stands to reason that MLM is important since it allows the model to adapt to new data-set specific tokenization.



Table \ref{tab:classifier_accuracy} also llustrates the impact of vocabulary extension. Indeed, reducing the vocabulary size from 16,000 to 8,000 tokens results in a minor accuracy decrease of 0.26\% for VIPI alone, while the MLM+VIPI approach experiences a more pronounced decline of 4.03\%. In contrast, increasing the vocabulary size to 32,000 tokens leads to a 1.24\% decrease in accuracy for VIPI alone, but a 2.16\% improvement with the MLM+VIPI method. This trend continues at 64,000 tokens, where VIPI alone decreases by 3.45\%, whereas the MLM+VIPI approach results in a 2.51\% improvement.

These findings suggest that expanding vocabulary size enhances model performance. This indicates that MLM effectively prepares the model to leverage larger vocabularies in domain-specific tasks.


\begin{table*}[t]
\centering
\begin{tabular}{l|c|c}
\hline
Vocabulary size & Change of accuracy & Change of accuracy\\
 & transfer only & transfer and MLM \\
\hline
16000 $\rightarrow$ 8000 & \textbf{-0.26\%} & -4.03\% \\
16000 $\rightarrow$ 32000 & -1.24\% & +2.16\% \\
16000 $\rightarrow$ 64000 & -3.45\% & \textbf{+2.51\%} \\
\hline
\end{tabular}

\caption{Relative change in downstream classifier accuracy and the impact of corpus-specific tokenization on a Kaggle medical dataset compared to standard fine-tuning with 16,000 tokens.}
\label{tab:classifier_accuracy}
\end{table*}

\subsection{Vocabulary Size and Inference time} \label{sec:sm}

In the medical domain, inference time might be critically important as it directly influences the speed and efficiency of healthcare analysis. Since patient data is highly sensitive and medical emergencies might occur in various conditions, local inference on diagnostic device rather than server-based inference might be vastly beneficial. This makes inference speed and efficiency paramount. Rapid inference facilitates real-time decision-making, swift processing of medical data, and timely responses in emergencies. Moreover, efficient inference optimizes resource utilization, accelerates patient care workflows, and enhances the overall experience for healthcare professionals. Faster inference times contribute to a more responsive, accessible, and effective healthcare system, ultimately improving patient care and outcomes, particularly in time-sensitive medical analyses.

\begin{figure}[h]
		\centering
    	\includegraphics[scale=0.5]{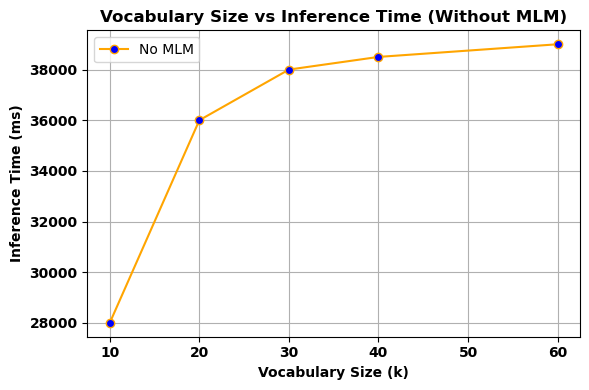}    
    	\caption{Change of classifier accuracy on Kaggle Medical dataset, inference time with respect to vocabulary size + VIPI only. }
    	\label{fig:inf_time}
\end{figure}

\begin{figure}[h]
		\centering
    	\includegraphics[scale=0.5]{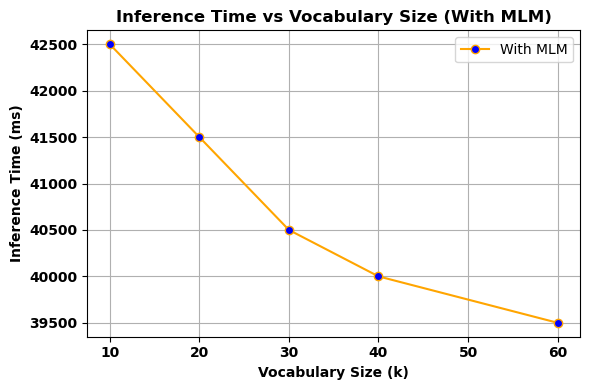}    
    	\caption{Relative change in accuracy of downstream classifiers on Kaggle Medical dataset, inference time with respect to vocabulary size after MLM and VIPI}
    	\label{fig:mlm_inf_time}
\end{figure}

Vocabulary size has a direct impact on inference time. In our experiment, as shown in Figure \ref{fig:inf_time}, we found that using a classifier with a larger vocabulary leads to increased inference time if the intermediary MLM step was not performed. However, as illustrated in Figure \ref{fig:mlm_inf_time}, incorporating a masked language model (MLM) training after vocabulary transfer results in decreased inference time as the vocabulary size increases.While larger vocabulary sizes generally increases inference time incorporating an MLM step allows the efficient handling of domain-specific vocabularies.This optimization reduces the computational burden during tokenization. Naturally, MLM enhances compression enabling the moder to process domain-specific data more efficiently throughout the all stages such as tokenization, embedding and final prediction generation.


\section{Discussion}

Several key aspects of vocabulary transfer for medical texts are evident from our analysis. First, the intermediary MLM step proves beneficial. Our further experiments show that even when applied to pre-existing tokenization, MLM on downstream data before training the classifier provides certain benefits. Likely, tokens rare in standard English but essential for medical NLP are better adjusted during this intermediate MLM phase. Additionally, the use of VIPI in conjunction with MLM appears to contribute to increased classifier accuracy especially if the vocabulary is extended.

Our findings also shows that the MLM step not only improves the model's classification performance but also leads to a reduction in inference time. By this we could say that MLM step compresses the token representation and optimizes the model for faster interference, even when the vocabulary size is significantly increased. By optimizing inference time, we can develop more efficient and responsive models. While more advanced and robust procedures may exist, our findings suggest that even a straightforward approach to vocabulary transfer in medical NLP can be significantly enhanced by expanding the vocabulary size.






\section{Conclusion}


This paper demonstrates the potential benefits of vocabulary transfer in medical natural language processing. We analyze the impact of various stages of vocabulary transfer on classification performance using medical datasets. Our findings indicate that increasing the vocabulary size leads to improved model performance.

\section{Limitations}
Our experiments are limited to text classification tasks using the base version of the BERT model tested on several specific datasets. We do anticipate that the proposed approach could be beneficial for other models and subdomains, since vocabulary transfer seem to have demonstrably similar effects in various domains, see \cite{yamshchikov2022bert, alexandrov2024mitigating, gee2022fast}. However, when working with models of bigger scale then BERT the effects of vocabulary transfer might be more or less pronounced.

\section*{Ethics Statement}
This paper complies with the \href{https://www.aclweb.org/portal/content/acl-code-ethics}{ACL Ethics Policy}.

\section*{Acknowledgements}
We would like to thank Mr. Pavel Chizhov and Mr. Alexey Tikhonov for their advice, productive ideas and support.

\bibliography{anthology}
\bibliographystyle{acl_natbib}

\end{document}